\pdfoutput=1

\documentclass[11pt, table]{article}

\usepackage[preprint]{acl}
\usepackage{xcolor}

\usepackage{times}
\usepackage{latexsym}

\usepackage[T1]{fontenc}

\usepackage[utf8]{inputenc}

\usepackage{microtype}

\usepackage{inconsolata}

\usepackage{graphicx}

\usepackage{algorithm}
\usepackage{algorithmic}
\usepackage{amsmath}
\usepackage{multirow}

\title{CuriousLLM: Elevating Multi-Document Question Answering with
LLM-Enhanced Knowledge Graph Reasoning}

\author{
 \textbf{Zukang Yang\textsuperscript{1}},
 \textbf{Zixuan Zhu\textsuperscript{1}},
 \textbf{Xuan Zhu\textsuperscript{1}},
\\
\\
 \textsuperscript{1}School of Information, University of California, Berkeley
\\
 \small{
   \textbf{Correspondence:} \href{mailto:zukangy, zzhu248, zhuxuan@berkeley.edu}{zukangy, zzhu248, zhuxuan@berkeley.edu}
 }
}

\begin{document}
\maketitle
\begin{abstract}
Large Language Models (LLMs) have achieved significant success in open-domain question answering. However, they continue to face challenges such as hallucinations and knowledge cutoffs. These issues can be mitigated through in-context learning by providing LLMs with relevant context before generating answers. Recent literature proposes Knowledge Graph Prompting (KGP) which integrates knowledge graphs with an LLM-based traversal agent to substantially enhance document retrieval quality. 
However, KGP requires costly fine-tuning with large datasets and remains prone to hallucination. 
In this paper, we propose CuriousLLM, an enhancement that integrates a curiosity-driven reasoning mechanism into an LLM agent. This mechanism enables the agent to generate relevant follow-up questions, thereby guiding the information retrieval process more efficiently.
Central to our approach is the development of the new Follow-upQA dataset, which includes questions and supporting evidence as input, with follow-up questions serving as ground truths. 
These follow-up questions either inquire about what is still missing to fully answer the user's query or use special tokens to signify that the retrieved evidence is sufficient. 
Our experiments show that CuriousLLM significantly boosts LLM performance in multi-document question answering (MD-QA), circumventing the substantial computational costs and latency from the original KGP framework. Source code: \url{https://github.com/zukangy/KGP-CuriousLLM}.
\end{abstract}

\section{Introduction}
\label{sec:intro}

\begin{figure*}[htbp]
    \centering
    \includegraphics[width=14cm, height=6cm]{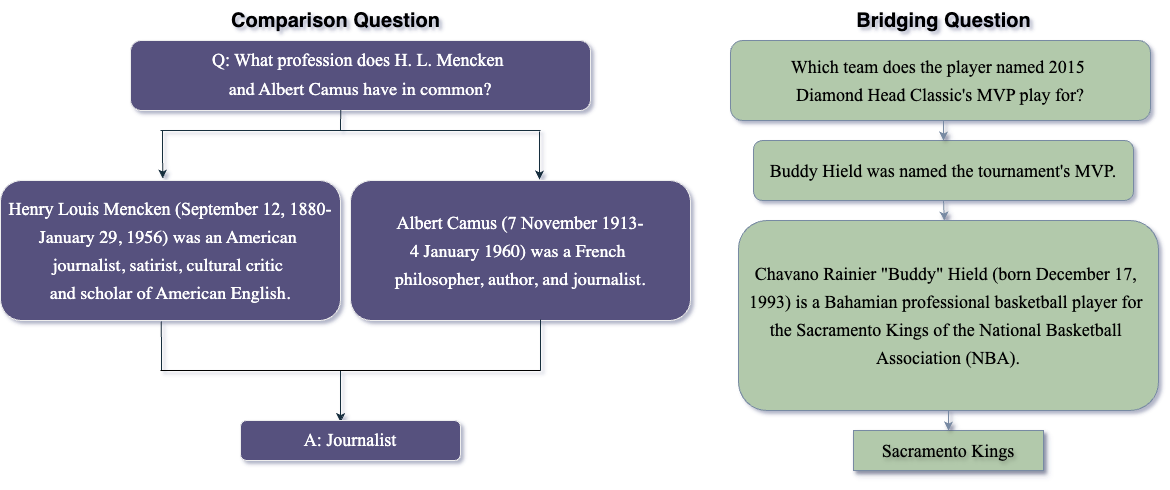}
    \caption{Two common types of questions in HotpotQA \cite{yang2018hotpotqa}: (1) Comparison questions require parallel reasoning over different documents. (2) Bridging questions require sequential reasoning.}
    \label{fig:questions}
\end{figure*}

Large Language Models (LLMs) have demonstrated remarkable success in open-domain question answering. However, they continue to face challenges such as hallucinations and knowledge cutoffs \cite{ji2023survey, ji-etal-2023-towards, yao2023llm, xu2024hallucination, wei2024measuring}. To address these issues, recent research has explored in-context learning with LLM, using external knowledge sources such as retrieved documents or knowledge graphs (KG) to improve their accuracy and reasoning ability \cite{Hogan_2021, Pan_2024, agrawal2024knowledge, shen2020exploiting, zhang2019ernie, rosset2021knowledgeaware, zhang-etal-2020-pretrain, kumar2020building, zhu2023minigpt}. Among the popular approaches are RAG \cite{lewis2020retrieval} and KAPING \cite{baek2023knowledgeaugmented}, which provide context by retrieving relevant documents or KG triplets to support the LLM reasoning process. Although these methods have proven effective, few approaches have fully integrated LLMs into the retrieval process, leaving a gap in the ability to efficiently navigate and extract relevant information from vast knowledge sources.

Recently, \citet{wang2023knowledge} propose Knowledge Graph Prompting (KGP), which incorporates a fine-tuned LLM agent into the KG traversal process. This agent predicts missing evidence based on the initial query and the retrieved documents. The predictions are then used to identify relevant passages from neighboring nodes in the KG through similarity ranking. With this prompt reformulation approach, KGP achieves state-of-the-art results in several benchmarks for factual consistency. 

However, during our experiments, we notice that the original KGP technique involves fine-tuning a T5 \cite{2020t5} agent with a large corpus. Despite this, the agent's performance remains limited due to hallucination: While correctly predicts missing evidence, it often uses irrelevant or erroneous keywords, which obscure the search for the actual piece of missing evidence, as shown in Table \ref{tab:hallucination}. Additionally, this technique tends to exhaust its preset search budget because it lacks a mechanism to determine when to stop the search, even when the retrieved documents are sufficient to answer the question. The more passages supplied to the LLM for response generation, the higher the latency of the QA system, since the LLM must reason through all passages to arrive at an answer.

To address these challenges, we propose a novel framework by fine-tuning an LLM agent to emulate the curious nature of a human researcher. Instead of predicting missing evidence, our agent asks follow-up questions to more efficiently guide the search toward missing evidence. Compared to the original approach, our CuriousLLM: 1) requires significantly fewer training samples for fine-tuning, 2) markedly improves MD-QA performance, and 3) can terminate the search before exhausting the preset search budget, thereby reducing latency. Our contributions are as follows.
\begin{itemize}
    \item \textbf{Follow-upQA Dataset:} We introduce a new dataset specifically designed to train LLMs to generate pertinent follow-up questions that enhance the retrieval process within the KGP framework for MD-QA tasks. In addition, we offer this dataset as a benchmark to inspire further research.
    \item \textbf{CuriousLLM Agent:} We design an LLM agent to ask follow-up questions, thus improving the efficiency and precision of the KGP framework without requiring extensive fine-tuning. 
    \item \textbf{Experimental Validation:} We present comprehensive experimental results that demonstrate significant improvements in both the performance and efficiency of the KGP framework using our approach. Furthermore, an ablation study highlights the enhanced reasoning capabilities of our LLM agent, particularly after fine-tuning on the Follow-upQA dataset.
\end{itemize}

\begin{table*}[ht!]
\centering
\resizebox{0.99\textwidth}{!}{ 
\begin{tabular}{p{5cm}|p{12cm}}
\hline
\textbf{Question:} & Which
magazine was started first: Arthur’s Magazine or First for Women? \\ \hline
\textbf{Retrived Evidence:} & Arthur's Magazine (1844 - 1846) was an American literary periodical published in Philadelphia in the 19th century. \\ \hline
\textbf{Missing Evidence:} & First for Women is a woman's magazine published by Bauer Media Group in the USA. The magazine was started in 1989. \\ \hline
\textbf{T5 Prediction:} & The publication of a woman's magazine is in London from 1921 to 1927. \\ \hline
\textbf{CuriousLLM Prediction:} & When was First for Women Magazine first started? \\ \hline
\end{tabular}
} 
\caption{Instance of T5 hallucination during graph traversal. Due to the erroneous keywords, T5 fails to identify the missing evidence. However, CuriousLLM succeeds in finding the missing evidence.}
\label{tab:hallucination}
\end{table*}

\begin{figure*}[ht]
    \centering
    \includegraphics[width=16cm, height=6.5cm]{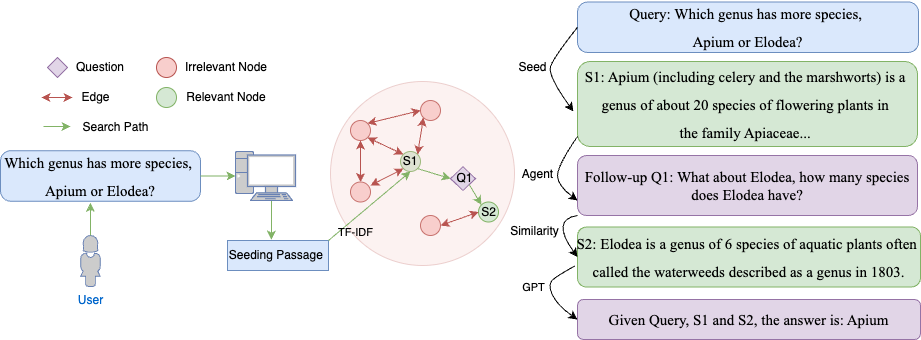}
    \caption{ Overview of the CuriousLLM workflow and an Follow-upQA example. Given a query, the system obtains seeding passages, and then starts searching for relevant documents; with follow-up question \textbf{Q1} 
    generated by the LLM agent, the unrelated passages S1 and S2 form a search path leading to the 
    final answer.}
    \label{fig:workflow}
\end{figure*}

\section{Methodology}
\label{sec:method}

\subsection{Follow-upQA Dataset}
We derive the new Follow-upQA dataset from the HotpotQA dataset \cite{yang2018hotpotqa}. HotpotQA is a multi-hop QA dataset containing questions and supporting passage pairs collected from Wikipedia. The questions in the dataset have three key features: 1) they cover a wide variety of topics; 2) they primarily consist of comparison and bridging questions (Figure \ref{fig:questions}), both of which are common in MD-QA tasks; and 3) they can be answered with at most two supporting passages \cite{xiong2021answering}. The first two features allow our LLM agent to learn from a diverse range of topics across both types of questions, while the third feature simplifies implementation and facilitates demonstration of Follow-upQA.

We create Follow-upQA through the following steps: First, we randomly sample questions from HotpotQA without replacement. Second, for the comparison questions, we randomly remove one of the two supporting passages. For the bridging questions, which require sequential reasoning, we retain only the first passage in the reasoning sequence. Next, we ask GPT-3.5 to generate a follow-up question based on the initial question and the supporting passage provided. Finally, if the selected question is a single-hop question, we prompt GPT-3.5-turbo \footnote{\url{https://platform.openai.com/docs/models/gpt-3-5}} to respond with "NA" to indicate that no further information is needed. In these steps, the follow-up question or the "NA" response serves as the ground truth for the question. We repeat this process until the budget is reached.

This process yields Follow-upQA, a dataset of  $\sim50K$ samples, with $59.5\%$ bridging questions, $26\%$ comparison questions, and $14.5\%$ "NA" or single-hop questions. 

\begin{table*}[thp!]
\centering
\resizebox{0.90\textwidth}{!}{ 
\begin{tabular}{p{0.48\textwidth}|p{0.48\textwidth}} 
\hline
\rowcolor[HTML]{B0C4DE} 
\textbf{Example 1} & \textbf{Example 2} \\ \hline

\textbf{(Input) Question:} & \textbf{(Input) Question:} \\ 
In what city is the university, for which Kim English played college basketball, located? & Tina Charles and Maya Moore were teammates on the UConn women's team that won championships in what years? \\ \hline

\textbf{(Input) Given:} & \textbf{(Input) Given:} \\ 
Kim English played college basketball for the University of Missouri before being selected by the Detroit Pistons with the 44th overall pick in the 2012 NBA draft. & In 2009 and 2010, Tina Charles and her teammate Maya Moore led the Connecticut Huskies to two undefeated national championships. \\ \hline

\textbf{(Ground Truth) Follow-up Question:} & \textbf{(Ground Truth) Follow-up Question:} \\ 
In which city is the University of Missouri located? & NA \\ \hline

\end{tabular}
} 
\caption{Follow-upQA examples of input questions, given information, and the corresponding follow-up questions generated.}
\label{tab:examples}
\end{table*}

In Table \ref{tab:examples}, we present examples of MD-QA tasks, illustrating the types of input questions and the corresponding given information, as well as the generated follow-up questions. These examples highlight how MD-QA systems, including our proposed model, are designed to handle complex questions that require reasoning across multiple pieces of evidence. For example 1, the follow-up question guides the graph traversal to look for information about University of Missouri's location. On the other hand, example 2 shows that the output is a special token "NA", signaling that sufficient information has been collected. 

\subsection{Knowledge Graph Construction}
Formally, we define a KG as $ G = (V, E, X)$, where $V = \{v_i\}_{i=1}^n$ denotes the set of nodes and $E \subset V \times V$ represents the relations between pairs of nodes. In our experimentation, each $v_i$ represents a passage and $X = \{x_i\}_{i=1}^n$ denotes a collection of dense representations with $x_i$ representing the passage embedding for $v_i$. 

In the original KGP experiments, constructing KG by multi-hop dense retriever (MDR-KG) \cite{wang2023knowledge} outperforms other KG construction approaches, such as k-nearest neighbors (KNN) \cite{Cunningham_2021} and TF-IDF. Following the methodology for building MDR-KG , we employ the MDR training technique \cite{xiong2021answering} to develop a BERT-based passage encoder using the HotpotQA dataset. This encoder is trained to predict subsequent supporting passages given an initial retrieved passage. The goal is to minimize the distances between passage pairs that are used together to answer questions in HotpotQA while simultaneously increasing the distances between unrelated passages through negative sampling. This training technique equips the encoder with reasoning capabilities, enabling it to understand the logical associations between different passages. Finally, we construct our KGs by encoding passages and connecting them based on cosine similarity.

\subsection{Curious LLM Traversal Agent}

We introduce CuriousLLM as our graph traversal agent to enhance the KGP framework. This approach is rooted in intuitive reasoning. For example, when asked to determine who is older, Bob Bryan or Mariaan de Swardt, and given information about Bob's age, one would intuitively ask a follow-up question about Mariaan's age. This follow-up question guides the search for relevant information. This intuition forms the basis of Follow-upQA and our model's training objective. The advantage of this approach is that, although the passages about Bob's and Mariaan's ages are unrelated, or, in other words, not semantically similar, the follow-up question about Mariaan's age creates a logical link between them. As a result, instead of matching two unrelated passages, once we find one passage, we can identify the other through follow-up questions. Furthermore, we train the LLM agent to know when to end the search, and this early termination mechanism significantly reduces the latency associated with the original T5 agent.

We use the Mistral-7B model \cite{jiang2023mistral} \footnote{\url{https://huggingface.co/mistralai/Mistral-7B-Instruct-v0.2}} as the backbone and fine-tune it on the Follow-upQA dataset with the objective of next-token prediction. Specifically, for each sample in the dataset, as shown in Table \ref{tab:examples}, we concatenate the question, the given passage, and the follow-up question. We employ QLoRA \cite{dettmers2023qlora} to train the model: we load the model at $8$ bit precision and then train a LoRA adapter using a LoRA rank of $32$, a learning rate of $10^{-5}$, and a batch size of $12$. The training is implemented with a split between the training validation test of $90\%$ - $5\%$ - $5\%$. Compared to the time required to train an identical T5 model in the original KGP framework, we reduce the training time of our model by $85\%$ using the same computing resources.

After generating a follow-up question at the current node in the KG, the question is compared against the neighboring passages of the node. We also employ a pre-trained Multi-QA sentence transformer \footnote{\url{https://huggingface.co/sentence-transformers/multi-qa-MiniLM-L6-cos-v1}}, which produces dense representations to minimize the semantic distance between the follow-up question and relevant passages. The search through KG is carried out using a breadth-first search strategy (BFS), as shown in Algorithm \ref{alg:KG_traversal}. This iterative process continues until reaching a predefined budget or the model deems it has sufficient information to answer the query.

Mathematically, given a user query $q_0$, we obtain a set of seeding passages $v_{j} \subset{\mathcal{V}^s}$ with TF-IDF. The agent accepts $q_0$ and the $j$-th seeding passage, and then generates a follow-up question $q_{1}^j$.  Formally,
\begin{equation}
q_{h}^j = \underset{v \in N_j}{\mathrm{arg\,max}}\; H(q_0, \|_{k=0}^{j}X_k)
\end{equation}
where $\|_{k=0}^{j}X_k$ concatenates the retrieved passages from the visited nodes on the same search path till the current node $v_j$. The choice of $H$ is a language model for next-token prediction. Moreover, the next passage $s_{j+1}$ is obtained as follows:
\begin{equation}
s_{j+1} = \underset{v \in N_j}{\mathrm{arg\,max}}\;\phi(g(q_{h}^j), g(X_{n})))
\end{equation} 
where $g$ is a sentence transformer, $X_n$ is passages from all neighboring nodes of node $v_j$ and $\phi$ is any similarity functions. See Algorithm \ref{alg:Follow-up QA Generation}. 

\subsection{LLM Response Generation}
After gathering sufficient evidence, we  leverage LLMs' capabilities  to provide a human-readable response to the user's query. We applied prompt engineering to guide the GPT4o-mini \footnote{\url{https://openai.com/index/gpt-4o-mini-advancing-cost-efficient-intelligence/}}, using the accumulated facts to generate an informed and coherent answer.

\begin{table*}[ht]
\centering
\begin{tabular}{c | c | c | c | c | c | c | c | c }
 \hline
 \multirow{2}{*}{Method} & \multicolumn{2}{c|}{HotpotQA} & \multicolumn{2}{c|}{2WikiMQA} & \multicolumn{2}{c|}{IIRC} & \multicolumn{2}{c}{MuSiQue} \\
 \cline{2-9}
 & Acc & EM & Acc & EM & Acc & EM & Acc & EM \\
 \hline
 None & 38.20 & - & 30.20 & - & 21.94 & - & 22.80 & - \\
 \hline
 TF-IDF & 67.40 & 45.60 & 42.80 & 47.66 & 28.09 & 28.27 & 28.60 & \underline{46.77} \\
 BM25 & 64.80 & 42.60 & 42.20 & 46.84 & 28.72 & 28.27 & 28.60 & 44.51 \\
 MDR & 70.60 & 50.12 & 44.20 & 45.38 & 31.45 & 30.96 & 32.80 & \textbf{47.59} \\
 \hline
 KGP\_T5 & 71.60 & \underline{51.77} & \underline{46.80} & \underline{49.55} & 33.30 & 31.57 & 34.00 & 45.85 \\
  \textbf{KGP\_Mistral\_ET (Ours)} & \underline{72.20} & 51.27 & 46.00 & 48.25 & \underline{34.80} & \underline{31.67} & \underline{34.80} & 45.85 \\
 \textbf{KGP\_Mistral (Ours)} & \textbf{73.80} & \textbf{52.42} & \textbf{49.40} & \textbf{50.29} & \textbf{36.69} & \textbf{32.07} & \textbf{36.50} & 45.95 \\
 \hline
 Golden & 81.40 & 100.00 & 69.80 & 100.00 & 64.57 & 100.00 & 53.80 & 100.00 \\
 \hline
\end{tabular}
\caption{Performance (\%) on $4$ multi-document question answering (MD-QA) benchmark datasets. KGP\_Mistral is our method. KGP\_Mistral\_ET is the version of our method with early termination. KGP\_T5 is a strong baseline. None: no passages but only the question is provided. Golden: supporting facts are provided along with the question. The best and run-up scores are in \textbf{bold} and \underline{underlined}.}
\label{tab:md-qa}
\end{table*}

\begin{figure*}[ht]
    \centering
\includegraphics[width=15.5cm, height=10.5cm]{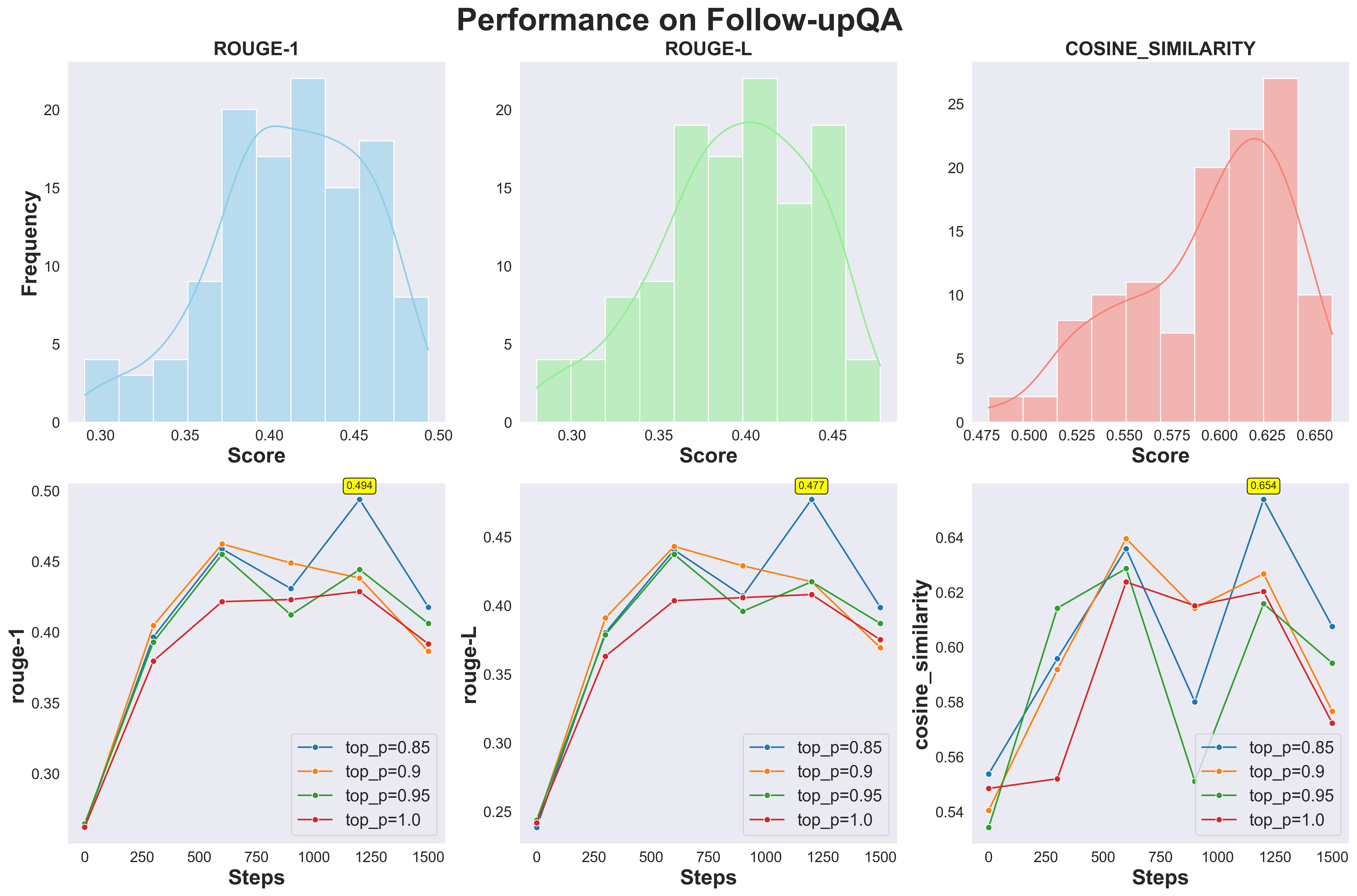}
    \caption{Benchmark Mistral-7B for Follow-upQA. First row: distribution plots for ROUGE-1, ROUGE-L, and cosine similarity across hyper-parameters. Second row: Mistral-7B performance at different training checkpoints.}
    \label{fig:performance}
\end{figure*}

\section{Experiments}
\label{Experiments}
To evaluate the multi-document question answering (MD-QA) capabilities of our CuriousLLM agent, KGP-Mistral, we adopt the experimental setup used by KGP-T5 \cite{wang2023knowledge}. In the original KGP-T5 evaluation, four MD-QA validation sets are used, each comprising $500$ questions sampled from the following datasets: HotpotQA, 2WikiMQA \cite{xanh2020_2wikimultihop}, IIRC \cite{ferguson2020iirc}, and MuSiQue \cite{trivedi2022musiquemultihopquestionssinglehop}. These datasets include $270K$, $120K$, $470K$, and $173K$ unique passages, respectively, which serve as supporting evidence or distracting passages.

Following our approach to KG construction, we treat each of these passages as a distinct node within the KG. For this evaluation, we limit the search scope to a 2-hop traversal, meaning that the system can retrieve and consider information from up to two edges away from the starting node in the graph. To ensure a fair comparison with KGP-T5, we standardize the retrieval parameters by selecting the top $30$ passages based on a similarity search to generate the answers.

\subsection{Evaluation Metrics}
We employ two evaluation metrics to assess the performance of our MD-QA system. First, we use accuracy (Acc) to measure the proportion of correctly answered questions. To evaluate accuracy, we prompt GPT-4o mini to compare each predicted answer with its corresponding ground truth. Second, we use exact match (EM) to assess the accuracy of information retrieval by calculating the proportion of facts correctly identified by the retriever compared to a set of golden facts. In particular, the EM implementation \footnote{\url{https://github.com/facebookresearch/multihop_dense_retrieval}} introduced by \citet{xiong2021answering} compares retrieved passages to their golden references token by token. However, in our experiments, we observe that passages from the validation sets do not always perfectly align with their golden counterparts due to the chunking strategy with overlaps used in the KGP-T5 experiments. This misalignment could potentially lead to an underestimation of the true EM scores. To address this issue, we opt to match passages based on cosine similarity instead. Our analyses indicate that while most exact matches exhibit similarity scores around $0.99$, some pairs with scores as low as $0.9$ are also considered equivalent.

\subsection{Performance and Analysis}
We study the MD-QA capability of our CuriousLLM agent (KGP-Mistral) employing several retrieval techniques as traversal agents for comparison, including the classic keyword-based methods TF-IDF and BM25, the encoder-based MDR, and a strong baseline, KGP-T5. Additionally, we include the "Golden" method, where the response generation LLM is provided with golden supporting passages, and the "None" method, where only the questions are given. As shown in Table \ref{tab:md-qa}, KGP-Mistral consistently achieves the highest performance across all benchmark datasets, with accuracy improvements of up to 9\% compared to BM25 on HotpotQA. Furthermore, compared to the original KGP framework (KGP-T5), KGP-Mistral delivers consistent accuracy gains across all datasets, averaging a 3\% improvement overall.

Keyword-based approaches such as TF-IDF and BM25 lack the reasoning capabilities necessary for complex MD-QA tasks. These methods primarily focus on retrieving passages that share keywords with the query or retrieved documents, without the ability to establish logical connections between different pieces of information. This limitation reduces their effectiveness in answering questions that require synthesizing information from multiple sources, as reflected in their lower accuracy scores compared to the LLM-based methods. The MDR method offers a more sophisticated approach by training a sentence encoder to bring passages used to answer questions closer in semantic space. This technique enhances the retrieval of related passages that are likely part of the evidence pair needed to answer the question. However, the LLM-based method, which builds on the MDR approach by incorporating a reasoning-driven LLM agent, shows better performance across all datasets. Furthermore, we show the impact of early traversal termination in Section \ref{early trans}.

In summary, our KGP-Mistral outperforms all other methods in accuracy and efficiency, establishing itself as a robust solution for multi-document question answering.

\section{Follow-upQA Benchmark}

\subsection{Evaluation on Follow-upQA}
We evaluate the fine-tuned Mistral-7B model on the Follow-upQA test set. We apply a train-validation-test split of $90\%$-$5\%$-$5\%$, resulting in $2.5K$ samples in the test set. The model is trained for $1,500$ steps, with a checkpoint saved every $300$ steps. In addition, we conduct a grid search for various decoding parameters, including temperature, top p, and maximum token length.

Figure \ref{fig:performance} presents histograms and line plots for ROUGE-1, ROUGE-L, and cosine similarity scores evaluating Mistral-7B on Follow-upQA. Most ROUGE scores concentrate around $0.4$, with a range of approximately $0.3$ to $0.5$. The cosine similarity scores range primarily from $0.475$ to $0.65$, indicating varying degrees of semantic alignment. The line plots reveal the performance of Mistral-7B at different checkpoints and highlight the impact of decoding parameters on the model's performance. The models achieve optimal performance at the $600$-step mark, with peak performance at step $1,200$. Specifically, the highest ROUGE-1 score is $0.494$ (top\_p=$0.85$). The best ROUGE-L score is $0.477$, achieved under the same conditions. For cosine similarity, the highest score is $0.654$, indicating close semantic alignment between the generated and golden questions.

Furthermore, the line graphs in Figure \ref{fig:performance} reveal that the initial performance of the raw Mistral-7B model is the lowest across all metrics, indicating a significant improvement through training. The gradual increase in scores demonstrates the model's enhanced ability to generate more accurate follow-up questions as training progresses. The peak performances annotated in the plots are achieved using the same model configuration: Mistral-7B at step $1,200$, with a temperature of $0.6$, top-p of $0.85$, and a maximum token length of $50$. Consequently, this model is selected for the MD-QA experiments. For a detailed analysis of Mistral-7B's performance across different training checkpoints, refer to the ablation study in Section \ref{ablation}.

\section{Conclusion}
\label{sec:concludes}
Our new CuriousLLM-enhanced KGP framework represents a significant advance in the field of MD-QA by integrating an LLM-guided prompt reformulation mechanism into the KG traversal process. By introducing the Follow-upQA dataset and a curiosity-driven LLM traversal agent, the system effectively addresses challenges like hallucination, inefficient retrieval, and the limitations of the original KGP framework. Extensive experiments show that this approach enhances MD-QA accuracy and efficiency while reducing computational overhead, making it a practical solution for real-world applications. This work paves the way for future research into optimizing LLM-guided retrieval processes by offering a scalable, robust framework balancing performance and resource efficiency.

\bibliography{main}

\cleardoublepage
\newpage

\appendix

\section{Related Work}

In the field of multi-document question answering (MD-QA), there has been significant progress in developing models that can efficiently retrieve and generate relevant information. MD-QA systems face unique challenges, as they require the model to process and integrate information from multiple documents while maintaining coherence and accuracy in the generated response. In this section, we provide an overview of the existing MD-QA approaches, categorized into three main types: retrieval-based models, generative models, and hybrid models that combine the strengths of both retrieval and generation. Each of these approaches brings distinct advantages and limitations to the task of answering complex questions that require reasoning over multiple documents.

\paragraph{Retrieval-based Models.}~Current retrieval-based models, such as TF-IDF \cite{Ramos2003UsingTT} and BM25 \cite{robertson2009probabilistic}, employ a term-document relevance mechanism to retrieve information based on lexical similarity to the query. Although these models perform well for questions that share explicit keywords with target documents, they often struggle when the query requires a deeper semantic understanding of the context \cite{lan2022research, viji2023hybrid, 10465440, 10009824}. To bridge this gap, encoder-based techniques, such as RNN encoders \cite{das2019multistep, schmidt2019recurrent, liu2019generative} and BERT-based encoders \cite{karpukhin2020dense, devlin2019bert, laskar2020utilizing}, leverage the power of deep learning to capture semantic information in texts. However, addressing the complexities of MD-QA presents additional challenges.

\paragraph{Generative Models.}~Recent advancements in LLMs have allowed models such as GPT \cite{brown2020language}, Llama \cite{touvron2023llama}, and Mistral \cite{jiang2023mistral} to provide fluent responses to user queries. These models are trained in vast corpora and further enhanced through Reinforcement Learning (RL) \cite{ziegler2020finetuning, rafailov2023direct} to effortlessly compose responses that mimic human conversations. However, the time and financial burdens associated with training, hosting, and maintaining an LLM are beyond the reach of many. In addition, LLMs are subject to issues like hallucination and knowledge cut-offs, limiting their effectiveness in the MD-QA domain.

\paragraph{Hybrid Models.}~Hybrid models represent a fusion of retrieval-based and generative approaches, equipping LLMs with a document retrieval system to provide relevant contextual information for response generation. This fusion effectively addresses the common issues faced by LLMs. Popular examples of such hybrid models include RAG and KAPING. Furthermore, \citep{Pan_2024, agrawal2024knowledge} summarize various strategies for unifying KGs and LLMs, including KG-enhanced LLMs \cite{shen2020exploiting, zhang2019ernie, rosset2021knowledgeaware}, LLM-augmented KGs \cite{zhang-etal-2020-pretrain, xie2022discrimination, Xie_2022, 9231227} and synergized LLM + KGs \cite{zhu2023minigpt, zhu2023minigpt4, thoppilan2022lamda, pub:30774}. These approaches significantly enhance the QA capabilities of LLMs.

\section{Algorithms}

In this section, we present two key algorithms designed to improve the performance of our CuriousLLM agent in the MD-QA framework. Algorithm 1 focuses on leveraging a KG to traverse and retrieve relevant information given a content-based user query. This process ensures that the model accesses the most pertinent context to answer complex questions. Algorithm 2 generates follow-up questions based on the retrieved passages to iteratively refine the search, enhancing the model's ability to gather more specific information. Together, these algorithms form the core of our system, enabling efficient and accurate responses in MD-QA tasks.

\begin{algorithm*}[hp]
\caption{LLM-based KG Traversal Algorithm for Retrieving Relevant Context Given a Content-based User Query}
\label{alg:KG_traversal}
\begin{algorithmic}[1]
\REQUIRE An initial query $q$ over a set of documents $\mathcal{D}$, the constructed KG $G = \{\mathcal{V}, \mathcal{E}, \mathcal{X}\}$ over $\mathcal{D}$, the LLM graph traversal agent $F_{agent}$, the preset passage budget $K$, the TF-IDF seeding passage retriever $g$
\STATE \textbf{Initialize seed passages $\mathcal{V}^s = g(\mathcal{V}, \mathcal{X}, q)$}
\STATE \textbf{Initialize the retrieved passage queue $\mathcal{P} = \{[v_i] \mid v_i \in \mathcal{V}^s\}$}
\STATE \textbf{Initialize the candidate neighbor queue $\mathcal{C} = [\mathcal{N}_i \mid v_i \in \mathcal{V}^s]$}
\STATE \textbf{Initialize the retrieved passage counter $k = \sum_{P_i \in \mathcal{P}} |P_i|$}
\WHILE {\textbf{queue $\mathcal{P}$ and queue $\mathcal{C}$ are not empty}}
    \STATE \textbf{$P_i \leftarrow \mathcal{P}.\text{dequeue}()$, $C_i \leftarrow \mathcal{C}.\text{dequeue}()$}
    \STATE \textbf{$\mathcal{V}_i' = \text{Traversal Agent}(q, P_i, C_i)$}
    \IF {\textbf{$\mathcal{V}_i' = \emptyset$}}
        \STATE \textbf{Terminate the loop}
    \ENDIF
    \FOR {\textbf{each $v \in \mathcal{V}_i'$}}
        \STATE \textbf{$\mathcal{P}.\text{enqueue}(P_i \cup \{v\})$, $\mathcal{C}.\text{enqueue}(\mathcal{N}_v)$}
        \STATE \textbf{$k \leftarrow k + 1$}
        \IF {\textbf{$k > K$}}
            \STATE \textbf{Terminate the loop}
        \ENDIF
    \ENDFOR
\ENDWHILE
\RETURN \textbf{Retrieved Passage Queue $\mathcal{P}$}
\end{algorithmic}
\end{algorithm*}

\begin{algorithm*}[hp]
\caption{CuriousLLM to Ask Follow-up Questions to Guide the Search}
\label{alg:Follow-up QA Generation}
\begin{algorithmic}[1]
\REQUIRE $q$ as initial query, $P_i$ as list of retrieved passages, $C_i$ as list of neighbor passages (these are inputs to \text{Traversal Agent} from \textbf{Algorithm 1}), preset $top\_k$ context budget, CuriousLLM $\mathcal{LM}$, sentence transformer Emb, similarity function $f_{\text{sim}}$, and ranker $\mathcal{R}$.
\STATE $q_{\text{new}} \leftarrow \text{CONCAT}(\{q\}, P_i)$
\STATE $q_{\text{follow\_up}} \leftarrow \mathcal{LM}(q_{\text{new}})$ from Eq (1)
\IF {$q_{\text{follow\_up}} == \text{'NA'}$}
    \RETURN $\{\}$
\ELSE
    \STATE $\text{Sim\_scores} \leftarrow f_{\text{sim}}(\text{Emb}(q_{\text{follow\_up}}), \text{Emb}(C_i))$
    \STATE $\text{Candidates} \leftarrow \mathcal{R}(\text{Sim\_scores})$ from Eq. (2) \RETURN the $top\_k$ in \text{Candidates}
\ENDIF
\end{algorithmic}
\end{algorithm*}

\begin{figure*}[hp]
    \centering
    \includegraphics[width=16cm, height=6cm]{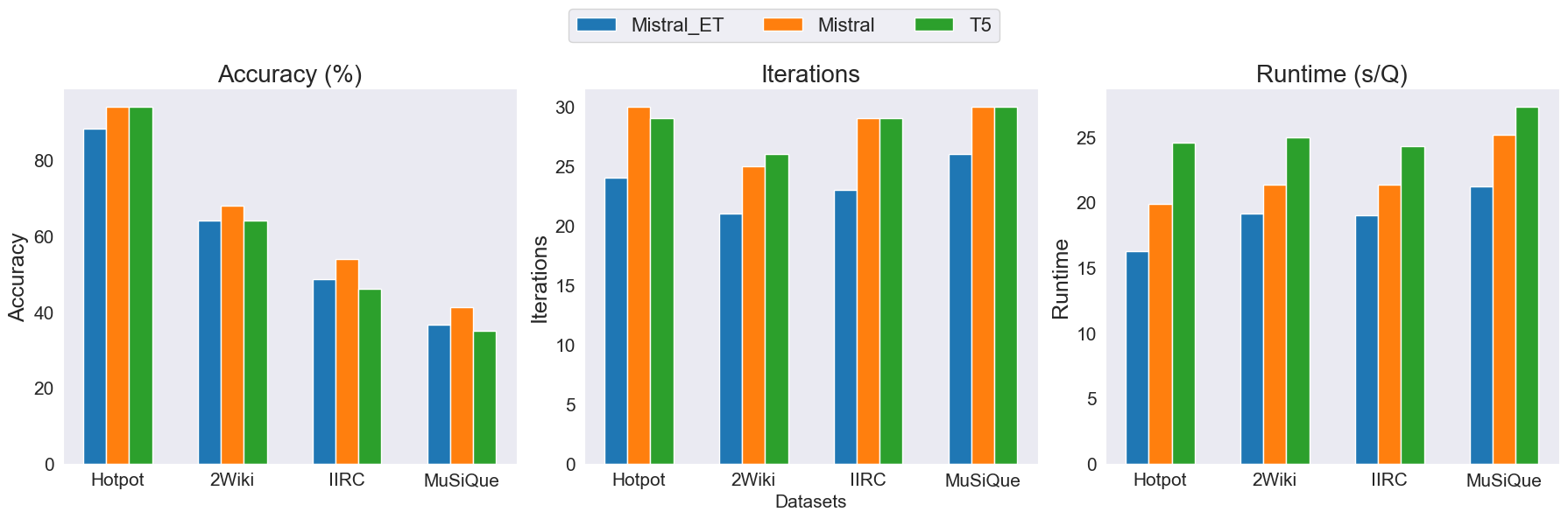}
    \caption{A comparison of MD-QA across LLM agents. Mistral\_ET is Mistral agent with early traversal termination. Accuracy calculates the correct rate of the questions that are early terminated by Mistral\_ET. Iterations can be interpreted as the number of nodes visited. Runtime records the average runtime in second per question.}
    \label{fig:early_stopping}
\end{figure*}

\begin{figure*}[tp]
    \centering
    \includegraphics[width=0.7\linewidth]{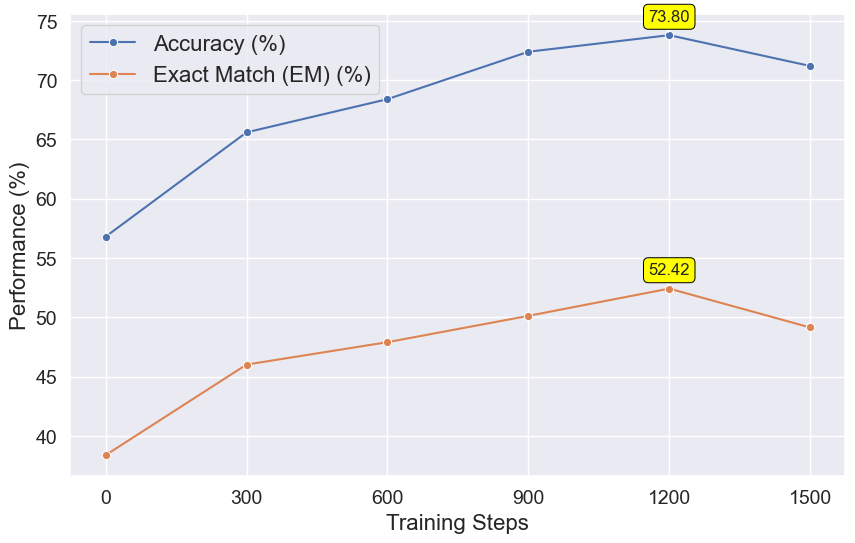}
    \caption{MD-QA performance on HotpotQA with Mistral agent at different training checkpoints.}
    \label{fig:Ablation_study}
\end{figure*}

\section{Impact of Early Traversal Termination} \label{early trans}

Our evaluation of early traversal termination using the CuriousLLM agent demonstrates substantial improvements in efficiency while maintaining competitive accuracy in MD-QA tasks. Our mistral model, fine-tuned to generate a termination signal ("NA") when it has gathered sufficient evidence, significantly reduces traversal time without compromising the quality of the final answer. This early termination feature allows the agent to stop the search process as soon as necessary information is identified, thus minimizing unnecessary computation and latency.

We compare three different agents: Mistral with early termination (Mistral-ET), the standard Mistral without early termination, and the T5 model. We collect questions that are closed early by Mistral-ET and evaluate the accuracy scores on these questions across all LLM agents. Moreover, we record the number of iterations or prompt reformulations for all questions, since fewer iterations typically result in lower latency. We also track the run-time across all questions.

The results in Table \ref{tab:md-qa} indicate that the early termination capability of Mistral-ET enables it to achieve accuracy levels similar to those of the T5 model, with fewer traversal iterations and reduced runtime. In this experiment, iterations refer to the number of nodes in the KG that need to be visited to gather sufficient evidence. Specifically, the accuracy of Mistral-ET matches closely that of T5, yet Mistral-ET consistently requires fewer node visits and less time per query, as illustrated in Figure \ref{fig:early_stopping}. Additionally, Mistral-ET shows only marginal differences in accuracy compared to its non-terminating counterpart, with the added benefit of faster execution. This efficiency gain is particularly relevant in real-world applications where quick response times are critical. Furthermore, since the experiments are conducted on a MacBook M2 Max, we anticipate even faster runtimes across all LLM agents with more powerful computing resources.

\section{Ablation Study} \label{ablation}

\paragraph{Training Step Analysis for MD-QA Optimization.}~We assess the MD-QA capabilities of the Mistral models at different end-to-end checkpoints on HotpotQA. In Figure \ref{fig:Ablation_study}, we observe a clear trend indicating that the MD-QA capability peaks at $1,200$ steps. At this checkpoint, the model achieves its highest accuracy and EM scores. This suggests that the model benefits significantly from training up to this point, with both accuracy and EM scores improving steadily from $0$ to $1,200$ steps. However, beyond $1,200$ steps, there is a slight decline in both metrics. Specifically, at $1,500$ steps, both the accuracy and EM scores decrease, suggesting that the model may begin to overfit the training data or that additional training steps introduce noise, slightly degrading performance.

Overall, the results underscore the importance of selecting an optimal number of training steps to maximize the performance of the KG traversal agent. Training up to $1,200$ steps appears to be the most effective strategy for this particular model and dataset, balancing sufficient learning with avoiding overfitting.

\section{Limitations}

While CuriousLLM demonstrates significant advancements in multi-document question answering (MD-QA), there are several limitations that merit discussion:

\paragraph{Question Scope.}~This system primarily focuses on addressing comparison and bridging questions, which require reasoning across multiple pieces of evidence. However, it leaves other common question types, such as what, where, and how, unexplored. Expanding the system to handle these broader types of questions would increase its applicability and robustness in diverse real-world scenarios.

\paragraph{Hardware and Scalability.}~The experiments conducted in this study used a single GPU, which limits the exploration of parallel processing and distributed computation. While the current setup validates the system's feasibility, its scalability to large-scale deployments in real-world environments will require more advanced hardware infrastructure and efficient parallelization techniques.

\end{document}